\documentclass[pdflatex,sn-mathphys-num]{sn-jnl}

\usepackage{graphicx}%
\usepackage{multirow}%
\usepackage{amsmath,amssymb,amsfonts}%
\usepackage{amsthm}%
\usepackage{mathrsfs}%
\usepackage[title]{appendix}%
\usepackage{xcolor}%
\usepackage{textcomp}%
\usepackage{manyfoot}%
\usepackage{booktabs}%
\usepackage{algorithm}%
\usepackage{algorithmicx}%
\usepackage{algpseudocode}%
\usepackage{listings}%

\theoremstyle{thmstyleone}%
\theoremstyle{thmstyletwo}%

\theoremstyle{thmstylethree}%

\begin{document}

\title[Article Title]{Privacy-Preserving Topic-wise Sentiment Analysis of the Iran–Israel–USA Conflict Using Federated Transformer Models}


\author[1]{\fnm{Md Saiful} \sur{Islam}}\email{2002047@icte.uftb.ac.bd }

\author*[2]{\fnm{Tanjim Taharat } \sur{Aurpa}}\email{aurpa0001@uftb.ac.bd}

\author*[2]{\fnm{Sharad} \sur{Hasan}}\email{sharad0001@uftb.ac.bd}

\author[3]{\fnm{Farzana} \sur{Akter}}\email{farzana0001@uftb.ac.bd}

\affil[1]{\orgdiv{Department of Educational Technology and Engineering}, \orgname{University of Frontier Technology, Bangladesh (UFTB), \city{Gazipur}, \country{Bangladesh}}}

\affil*[2]{\orgdiv{Department of Data Science and Engineering}, \orgname{University of Frontier Technology, Bangladesh (UFTB), \city{Gazipur}, \country{Bangladesh}}}

\affil[3]{\orgdiv{Department of IOT and Robotics Engineering}, \orgname{University of Frontier Technology, Bangladesh (UFTB), \city{Gazipur}, \country{Bangladesh}}}


\abstract{

The recent escalation of the Iran–Israel–USA conflict in 2026 has triggered widespread global discussions across social media platforms. As people increasingly use these platforms for expressing opinions, analyzing public sentiment from these discussions can provide valuable insights into global public perception. Understanding how public opinion evolves during international conflicts is important for researchers, policymakers, and media analysts.
This study aims to analyze global public sentiment regarding the Iran–Israel–USA conflict by mining user-generated comments from YouTube news channels. The work contributes to public opinion analysis by introducing a privacy-preserving framework that combines topic-wise sentiment analysis with modern deep learning techniques and Federated Learning.
To achieve this, approximately 19,000 YouTube comments were collected from major international news channels and preprocessed to remove noise and normalize text. Sentiment labels were initially generated using the VADER sentiment analyzer and later validated through manual inspection to improve reliability. Latent Dirichlet Allocation (LDA) was applied to identify key discussion topics related to the conflict. Several transformer-based models—including BERT, RoBERTa, XLNet, DistilBERT, ModernBERT, and ELECTRA—were fine-tuned for sentiment classification. The best-performing model was further integrated into a federated learning environment to enable distributed training by preserving user data privacy. Additionally, Explainable Artificial Intelligence (XAI) techniques using SHAP were applied to interpret model predictions and identify influential words affecting sentiment classification. Experimental results demonstrate that transformer models perform effectively and among them, ELECTRA achieved the best performance with 91.32\% accuracy. The federated learning also maintained strong performance while preserving privacy, achieving 89.59\% accuracy in a two-client configuration. Topic-wise sentiment analysis further revealed that discussions about geopolitical tensions and political controversies generated higher negative sentiment, while topics related to leadership, religion, and peace showed comparatively more positive reactions.

}




\maketitle

\section{Introduction}\label{sec1}
The rivalry between USA-Israel ally and Iran is a long-running issue. This has been escalated recently and immediately triggered as a war on 28th February 2026 with the launch of a large-scale air and missile strike on Iran by the USA-Israel alliance. This war has upended the geopolitical dynamics of the Middle East\cite{britannica2026iranconflict}. The effect of this war is not limited only to the Middle East or to the rival countries. Because of this era of globalization, any conflict like a war between some countries, has a grave impact on the whole world. This war can cause geopolitical instability with the involvement of great powers like Russia and China. Iran controls the Strait of Hormuz, through which about 20\% of the world’s oil passes. With the blocking of the Strait of Hormuz, oil prices could go rocket-high, which can crash the global market, causing a grave danger for the developing countries. The war can also cause a large-scale use of advanced weapons, including nuclear weapons. It can cause many civilian casualties, infrastructure collapse, polarization of the world, rise in global tensions, and acceleration of energy transition away from oil \cite{mfcnn_bilstm_ukraine_russia_sentiment}. 

Since the war has shaken the whole world, people from every corner of the world are expressing their opinions through social media. Worldwide, social media platforms have evolved into interactive spaces for individuals for open discussion and a strong way to express their voices. These platforms have become an essential part of daily life in modern societies, as they enable users to access vast amounts of information and freely share their thoughts and emotions. YouTube is a popular social media platform where people from around the world share their opinions in the comment sections of international news channels. The large volume of user-generated content provides valuable insights into how public opinion changes across different perspectives.

In our study, we have analyzed public sentiment on Iran vs USA-Israel war using data collected from YouTube comments posted on various well-known international news channels. These comments highlight the contrast between the objective tone of news reporting and the subjective reactions of the public \cite{islam2025wordsofwar}. This sentiment analysis is necessary to understand the global public opinion and reaction towards this war.  Since the public sentiment changes rapidly during these kinds of conflicts based on different incidents, it is crucial to analyze the sentiment to have a real time insight. Different media houses owned by different bodies can post video reports biased to their own perspective. Our study analyzes the sentiment under the videos which can provide an unbiased view of the actual scenario, which in turn can also facilitate the detection of misinformation and propaganda. This unbiased information is also helpful for news reports and articles. Analyzing the sentiment of this diverse data can help to enrich data-driven policy discussion and decision-making for countries and international organizations. This sentiment analysis can reveal collective emotions and psychological impacts during conflicts \cite{liyih2024sentiment}.
It is challenging to analyze sentiment on a multi-faced issue like war because the data in hand contains diverse perspectives. Because of the unstructured nature of the data, unsupervised sentiment analysis has been used. Rather than focusing solely on overall sentiment analysis, this study first identifies the key topics associated with the Iran–Israel–USA war from social media data, specifically YouTube comments. Subsequently, topic-wise sentiment analysis is performed to examine how public opinions vary across different aspects of the conflict.

The authors propose a privacy-preserving methodology that utilizes YouTube comments along with modern Transformer-based models and Federated Learning. This approach aims not only to predict public perspectives regarding the Iran–Israel–USA war but also to ensure the protection of users’ privacy while analyzing social media data. To the best of our knowledge, this study is the first work that focuses on public opinion mining and prediction related to the 2026 Iran war using a privacy-aware framework. The main contributions of our study are listed below:
\begin{itemize}
\item After labeling the sentiments using VADER (Valence Aware Dictionary and Sentiment Reasoner), we have manually validated the sentiment labels. This manual validation has increased the robustness and credibility of our data.
For the extraction of the main topics being discussed within the comments posted, the technique of Latent Dirichlet Allocation (LDA) was used. 
\item We present a detailed analysis of public opinion across different perspectives of the war using the labeled dataset. The analysis demonstrates how sentiment polarity varies across different topics related to the war, highlighting changes in public perception depending on the discussed aspec
\item We have considered several transformer models for sentiment classification by fine-tuning them using the labeled dataset. ELECTRA (Efficiently Learning an Encoder that Classifies Token Replacements Accurately) was selected as the final model for the topic wise sentiment classification that introduces a discriminative pretraining mechanism.
\item Our study has enhanced data security and privacy by employing ELECTRA in a Federated learning environment.
\item We have employed Explainable Artificial Intelligence (XAI) to improve the interpretability of the sentiment classification. SHAP method has been employed to evaluate the effects that individual words have on predictions.
\end{itemize}

\section{Literature Review} 
\subsection{Topic-wise Sentiment Analysis with Transformers}
Topic-wise sentiment analysis aims to identify sentiments expressed toward specific topics within a text rather than determining overall sentiment. This approach provides more detailed insights into opinions by associating sentiment with particular entities, attributes, or themes discussed in textual data. 
Authors in \cite{madhurika2025deep} proposed SentiNet, a hybrid deep learning architecture designed to improve sentiment classification of customer reviews. The model integrates CNNs, BiLSTM, and an attention mechanism to capture both local textual features and long-term contextual dependencies. Experiments conducted on benchmark datasets such as IMDb, Twitter, and Yelp demonstrated that the proposed model achieves high performance, with 98.68\% accuracy and a 97.48\% F1-score, outperforming traditional CNN, LSTM, and BiLSTM models. Albladi et al. proposed TWSSenti \cite{albladi2025twssenti}, a hybrid transformer framework for topic-wise sentiment analysis that combines transformer models such as BERT, GPT-2, RoBERTa, XLNet, and DistilBERT to improve robustness across both short, noisy text and longer reviews. The study used Sentiment140 and IMDB datasets and achieved about 94\% accuracy on Sentiment140 and 95\% on IMDB, outperforming several standalone transformer baselines. The paper shows that combining complementary transformers can improve generalization, reduce errors, and make sentiment systems more practical. 
Paper \cite{putri2026topic} examined topic-based sentiment analysis of Jamsostek Mobile (JMO) application reviews using a combination of IndoBERT for sentiment classification and BERTopic for topic discovery. Their dataset consisted of 2,846 Google Play reviews collected in 2025, and the study found that 75\% of the comments were negative, while the classifier reportedly achieved 100\% accuracy. This paper shows how sentiment classification and topic modeling can be integrated to reveal both polarity and the main sources of user dissatisfaction in a public-service mobile app. 
Authors in \cite{shafana2022does} studied public opinion on online learning in the South Asian region during COVID-19 using Twitter data, lexicon-based sentiment analysis, and LDA topic modeling. From 2,103 tweets filtered from a larger collection, they found that 63.2\% were positive, 30.5\% neutral, and about 6.3\% negative, suggesting that people generally accepted online education as a necessary solution during the pandemic. Their topic analysis showed that positive discussions often focused on affordability and access to courses, while negative discussions emphasized weak internet access, reduced classroom attention, and the loss of emotional connection in online classes.
Susan et al. conducted a longitudinal analysis of COVID-19 vaccine-related tweets in India, combining AFINN-based sentiment analysis with dynamic LDA topic modeling at points where sentiment sharply changed over time \cite{susan2025longitudinal}. Using 44,130 geo-located tweets, the study found that 51.38\% were neutral, 38.84\% positive, and 9.78\% negative. Rather than treating sentiment as static, the authors linked peaks in positive or negative sentiment to real events such as vaccine approvals, shortages, trial results, and policy announcements. 
Islam et al. proposed “Words of War”, a hybrid BERT–CNN framework for topic-wise sentiment analysis of public opinions on the Russia-Ukraine war using social media data \cite{islam2025words}. The study collected 85,904 YouTube comments from major news channels such as BBC, CNN, and Al-Jazeera between January 2022 and August 2024. First, unsupervised BERT-based topic modeling was used to identify ten major discussion themes, including geopolitics, civilians, war crimes, media influence, economy, and NATO involvement. Sentiment labels (positive, negative, neutral) were generated using the VADER lexicon-based sentiment analyzer. In this hybrid model, BERT captures contextual semantic information, and CNN extracts important local textual features, while topic information is incorporated as an additional input feature. Experimental results showed that the proposed model achieved 92.26\% accuracy on the YouTube dataset, 92.98\% on a Twitter dataset, and 88.87\% on a Reddit dataset, outperforming several baseline transformer models such as BERT, RoBERTa, XLNet, and ELECTRA.

\subsection{XAI for Topic wise Sentiment Analysis}
Explainable Artificial Intelligence (XAI) provides deeper insights into the factors influencing sentiment classifications, while the framework ensures strong prediction performance \cite{elbasiony2024xai}. XAI helps to gain a deeper understanding of machine learning models based on the factors that affect the predictions \cite{vaidya2025meta}. Recent research has utilized Explainable Artificial Intelligence (XAI) in Educational Data Analysis (EDA) to enhance the transparency of predictive models based on the significant features that affect the decisions \cite{almazroei2026enhancing}.

SHAP summary plots integrate feature importance and feature effects by representing each instance with a Shapley value, enabling visualization of how features influence model predictions \cite{saini2024grainy}. In another study, improved interpretability of sentiment prediction models helped businesses derive actionable insights from customer feedback \cite{rudro2026deepbert}. SHAP analysis shows how words represented as tokens influence model predictions, where tokens such as “wonderful” and “balanced” have a strong positive impact on the output \cite{le2026xai}. SHAP analysis highlights both positive and negative token contributions, where words such as “fresh” and “charm” contribute positively, while phrases like “a typical romantic lead” indicate negative sentiment \cite{thogesan2025integration}. In another study, a SHAP force plot revealed that keywords such as “notorious,” “bad,” “unbelievable,” and “hacked” strongly contributed to a highly negative review \cite{rizvi2025enhancing}. In another topic based analysis, SHAP used to interpret the word levels of significance on model classification \cite{shin2025explainable}.

\subsection{Federated Learning on Sentiment Analysis}
Federated learning(FL) is used in sentiment analysis to enable collaborative model training across distributed datasets while preserving privacy, enhancing security, and avoiding centralized data collection. Several works have been conducted in recent times utilizing this modern privacy-preserving technique. The modern FL paradigm is commonly traced to McMahan et al., who formalized decentralized training with iterative model averaging and showed that useful deep models could be learned from distributed, potentially non-IID client data \cite{mcmahan2017communication}. One example is the geographically distributed tweet-based study \cite{sethi2021semi}, which proposed a semi-supervised federated sentiment analysis pipeline across multiple locations. Their work is important because it framed sentiment data as naturally partitioned by geography and privacy constraints, and showed that federated training could be used to build a broader model without pooling raw tweets centrally. 
Qin et al. presented one of the clearest task-specific advances in this area with Improving Federated Learning for Aspect-based Sentiment Analysis via Topic Memories \cite{qin2021improving}. Their paper argued that ABSA in FL is especially difficult because each client sees only a limited local distribution, which weakens the model’s understanding of sentiment patterns and aspect context. To address this, they introduced topic memories to better capture shared semantic knowledge across clients while maintaining decentralized training.
Basu et al. studied the combined effects of differential privacy and federated learning for BERT-based NLP models on depression- and sexual-harassment-related tweets \cite{basu2021benchmarking}. 
Similarly, a recent paper \cite{ahsan2024privacy} on mental-health sentiment analysis proposed an FL-BERT framework combined with data obfuscation to balance predictive performance and privacy.
The paper \cite{gholamiangonabadi2024federated} investigates the challenges of applying federated learning to sentiment analysis when the client data distributions are non-IID (non-independent and identically distributed). The authors evaluate how different deep learning models behave under such heterogeneous data conditions and analyze the sensitivity of federated models to data imbalance. Their findings show that accuracy decreases as data imbalance increases.

\section{Methodology}\label{sec2}
In this section, the preliminaries of the proposed methodology and the overall research framework are described. The major part
of the methodology of this research is explained below.

\subsection{Data Acquisition}

To analyze public opinion regarding the discussed topic, a large number of user-generated comments were collected from YouTube.  Comments were collected from various well-known international news channels, including BBC, CNN, FOX News, WION, Sky News, Firstpost, and Al Jazeera. These news channels are known to frequently upload news-related videos on their channels, which are mostly viewed and discussed by the publics.

For the collection of comments, the time period chosen was between March 1 and March 7, during which comments were scraped using an automated tool from various videos uploaded by the aforementioned news channels. A total of approximately 19,000 comments were collected for the analysis. These comments contained various opinions and sentiments of the general public from different geographical locations and backgrounds
The comments collected were all publicly available and were stored in a structured format for further processing and analysis.

\subsection{Data Preprocessing}

The collected YouTube comments were preprocessed to remove noise and improve text quality before analysis \cite{haider2025comprehensive}. Several standard text cleaning steps were applied shown in  Figure\ref{fig:preprocessing_pipeline} :

\begin{figure}[htbp]
\centering
\includegraphics[width=0.95\textwidth]{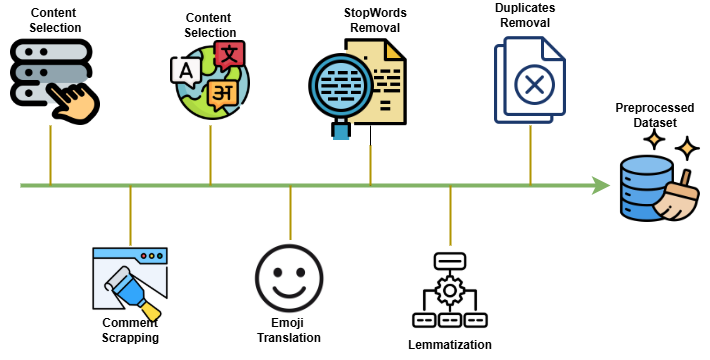}
\caption{Overview of the data preprocessing pipeline}
\label{fig:preprocessing_pipeline}
\end{figure}

\begin{itemize}
    \item \textbf{Language Translation:} Comments that were not in English were translated into English.
    
    \item \textbf{Emoji Translation:} Emojis were converted into textual descriptions to preserve emotional meaning \cite{apu2025explainable}.
    
    \item \textbf{Special Character Removal:} This includes the removal of URLs, numbers, punctuation, and other text symbols.
    
    \item \textbf{Stopword Removal:} Common English stopwords were removed while keeping important negation words.
    
    \item \textbf{Lemmatization:} Words were reduced to their base form to normalize variations.
    
    \item \textbf{Duplicate Removal:} Duplicate comments were removed to prevent redundancy in the dataset.
\end{itemize}

After preprocessing, the further analysis such as sentiment labeling and topic modeling .

\subsection{Data Labeling}

After preprocessing, comments were labeled to identify their sentiment polarity as well as underlying discussion topics. To perform this task, two techniques were employed: sentiment labeling using VADER and topic extraction using Latent Dirichlet Allocation (LDA).

\subsubsection{Sentiment Labeling}

For sentiment labeling, a sentiment analysis tool called VADER (Valence Aware Dictionary and Sentiment Reasoner) is employed \cite{nurcahyawati2025exceeding}. It is a lexicon and rule-based sentiment analysis tool commonly applied in social media data analysis tasks \cite{mehenaoui2024comparative}. VADER calculates a compound sentiment score for each comment, where the score ranges from $-1$ (extremely negative) to $+1$ (extremely positive).

Using this compound sentiment score, each comment in a social media data set is automatically labeled into one of three classes: \textit{positive}, \textit{negative}, and \textit{neutral}. Following the standard VADER threshold values, comments with a compound score greater than or equal to $0.05$ were labeled as positive, comments with a score less than or equal to $-0.05$ were labeled as negative, and comments with scores between $-0.05$ and $0.05$ were labeled as neutral. This automated labeling process allows efficient sentiment annotation of large-scale social media datasets shown in Algorithm \ref{alg:vader_sentiment}.

\begin{algorithm}[H]
\caption{Sentiment Labeling using VADER}
\label{alg:vader_sentiment}
\begin{algorithmic}[1]
\Require Preprocessed comment dataset $D$
\Ensure Sentiment label for each comment

\State Initialize the VADER sentiment analyzer
\For{each cleaned comment $c \in D$}
    \State Compute the compound sentiment score $s$ using VADER
    \If{$s \geq 0.05$}
        \State Assign sentiment label as \textbf{Positive}
    \ElsIf{$s \leq -0.05$}
        \State Assign sentiment label as \textbf{Negative}
    \Else
        \State Assign sentiment label as \textbf{Neutral}
    \EndIf
\EndFor
\State Store the sentiment label and compound score in the dataset
\end{algorithmic}
\end{algorithm}

\noindent\textbf{Notation:}

\begin{itemize}
\item $D$ : The set of preprocessed comments in the dataset.
\item $c$ : A single comment from the dataset.
\item $s$ : The compound sentiment score generated by VADER.
\item Positive : Comments with $s \geq 0.05$.
\item Negative : Comments with $s \leq -0.05$.
\item Neutral : Comments with $-0.05 < s < 0.05$.
\end{itemize}

\subsubsection{Topic Extraction using LDA}

For the extraction of the main topics being discussed within the comments posted, the technique of Latent Dirichlet Allocation (LDA) was used \cite{li2025shaping}. This technique of topic modeling uses the probabilistic model to identify the underlying hidden patterns within a group of documents \cite{jin2025research}. In this research work, the number of topics was set to 10 to identify the underlying discussion topics within the comments. Each topic was represented by a group of keywords that described the underlying theme of the comments.

\subsubsection{Manual Validation of Sentiment Labels}
To validate the accuracy of the automatically generated sentiment labels, a manual evaluation process was carried out \cite{islam2025wordsofwar}. A random sample of the dataset was chosen for cross-checking purposes. The sample contained approximately 20\% of the original dataset. For each of the randomly chosen comments, the original text along with the assigned sentiment labels by the VADER model was manually inspected.

The main aim of the manual evaluation process was to check whether the assigned sentiment category for the comments accurately represented the meaning of the comments. By carrying out the manual evaluation process, the accuracy of the automatically generated labels for the dataset was increased.

\subsection{Word Cloud Visualization}

To gain an intuitive understanding of the most frequently used words in the collected comments, a word cloud visualization technique was applied \cite{pham2025reddit}. Word clouds provide a visual representation of word frequency within a text corpus, where words appearing more frequently are displayed in larger font sizes \cite{gupta2025sentiment}.

\subsection{Transformer Model Selection}

To identify the most appropriate model for sentiment classification, several transformer models were considered in the current research. It is established that the transformer model can perform effectively in different natural language processing tasks.

For instance, several transformer models such as BERT, RoBERTa, XLNet, DistilBERT, ELECTRA, and ModernBERT were considered for sentiment classification in the current research. These models are also known as state of the art models \cite{aurpa2024instructnet, aurpa2021progressive}.
These models were chosen for the current research due to their popularity in natural language processing.

Each model was fine-tuned using the labeled dataset to classify comments. The objective of this step was to identify a robust model capable of effectively capturing sentiment patterns in social media text \cite{jeba2024facebook}.

\subsubsection{ELECTRA-based Sentiment Classification}

Among the evaluated transformer models, ELECTRA was selected as the final model for sentiment classification. ELECTRA (Efficiently Learning an Encoder that Classifies Token Replacements Accurately) is a transformer-based architecture that introduces a discriminative pretraining mechanism \cite{karayiugit2026electra}. The model is different from other transformer-based models in that it uses a discriminator to classify whether tokens in a given text sequence are replaced or not \cite{csenturk2026multi}. ELECTRA delivers fast, resource-efficient performance, which makes it well-suited for high-volume platforms \cite{philipo2026sentiment}.

This method enables ELECTRA to efficiently learn contextual information from text data \cite{amirthasaravanan2026enhanced}. The model was found to be effective in leveraging training data. The effectiveness of ELECTRA in learning contextual information from text data was found to be significant. The model was chosen as the primary model for sentiment classification due to its effectiveness in learning contextual information. The model was fine-tuned to classify the sentiment of comments in the dataset.

\begin{figure}[htbp]
\centering
\includegraphics[width=\textwidth]{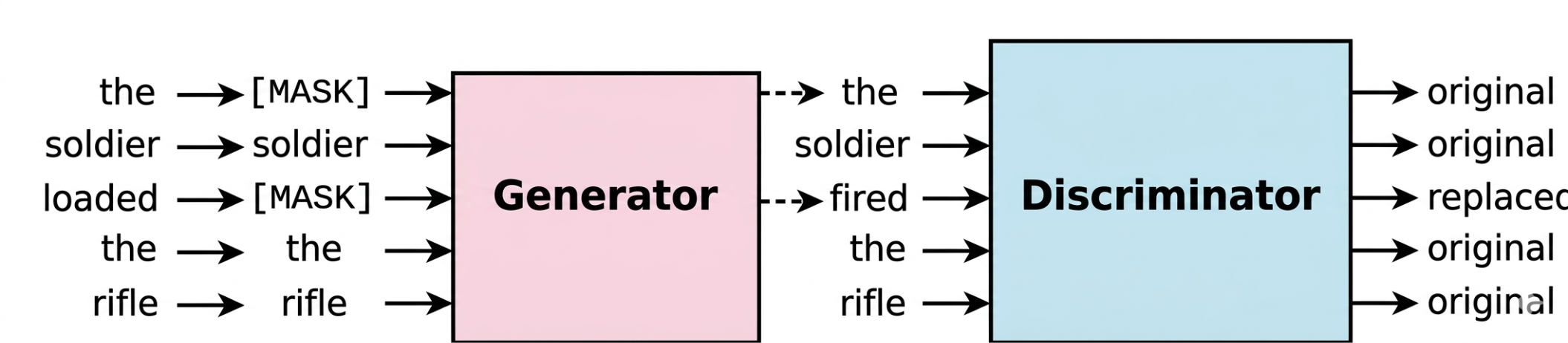}
\caption{Illustration of the ELECTRA generator and discriminator framework.}
\label{fig:electra_mechanism}
\end{figure}
Figure~\ref{fig:electra_mechanism} illustrates the core working mechanism of the ELECTRA model. The generator predicts replacements for masked tokens, while the discriminator learns to identify whether each token in the sequence is original or replaced, enabling efficient contextual representation learning.

\subsubsection{Federated ELECTRA}

In this study, a federated learning method was employed in training the ELECTRA model for sentiment classification. The cleaned YouTube comments, together with their corresponding topic keywords, were used as input features, while the sentiment labels obtained by VADER were used as target classes.  This combination allows the model to capture both textual sentiment information and thematic context extracted through LDA topic modeling.

For federated learning, dataset partitioning was employed, in which multiple clients were created with stratified partitioning, ensuring balanced sentiment classes in each client. Each client received the current global ELECTRA model (\texttt{google/electra-base-discriminator}) and performed local training using its own data. During preprocessing, comments and topic keywords were jointly tokenized with a maximum sequence length of 256.
 Federated Learning (FL) has recently attracted significant attention in Sentiment Analysis (SA) as it enables decentralized model training while preserving data privacy across multiple clients \cite{pandya2026pans}.
To preserve data privacy, the raw training data remained on the local clients and were never shared with the central server. Instead, only the locally updated model parameters were transmitted to the server. After local training, the updated parameters from each client were aggregated using the Federated Averaging (FedAvg) algorithm to update the global model. This iterative communication was performed for several rounds, enabling the global model to learn from the distributed data while preserving the data privacy.

The working procedure of the Federated ELECTRA model is presented in Algorithm~\ref{alg:federated_electra}.

\begin{algorithm}[H]
\caption{Federated ELECTRA for Sentiment Classification}
\label{alg:federated_electra}
\begin{algorithmic}[1]
\Require Dataset $D$, number of clients $N$, communication rounds $R$
\Ensure Trained global ELECTRA model $M_g$

\State Split dataset $D \rightarrow D_{train}, D_{val}$
\State Partition $D_{train} \rightarrow \{D_1, D_2, ..., D_N\}$
\State Initialize global ELECTRA model $M_g$

\For{$r = 1$ to $R$}
    \For{$i = 1$ to $N$}
        \State $M_i \leftarrow Train_{ELECTRA}(M_g, D_i)$
    \EndFor
    \State $M_g \leftarrow FedAvg(M_1, M_2, ..., M_N)$
\EndFor

\State Evaluate $M_g$ on validation set $D_{val}$
\end{algorithmic}
\end{algorithm}

\noindent\textbf{Notation:}
\begin{itemize}
\item $D$ : Preprocessed dataset.
\item $D_i$ : Local dataset of client $i$.
\item $N$ : Number of federated clients.
\item $R$ : Number of communication rounds.
\item $M_g$ : Global ELECTRA sentiment classification model.
\item $M_i$ : Local ELECTRA model trained on client $i$.
\item $Train_{ELECTRA}(\cdot)$ : Fine-tuning ELECTRA on local client data.
\item $FedAvg(\cdot)$ : Federated Averaging used to aggregate client models.
\end{itemize}

\begin{figure}[H]
\centering
\includegraphics[width=0.7\textwidth]{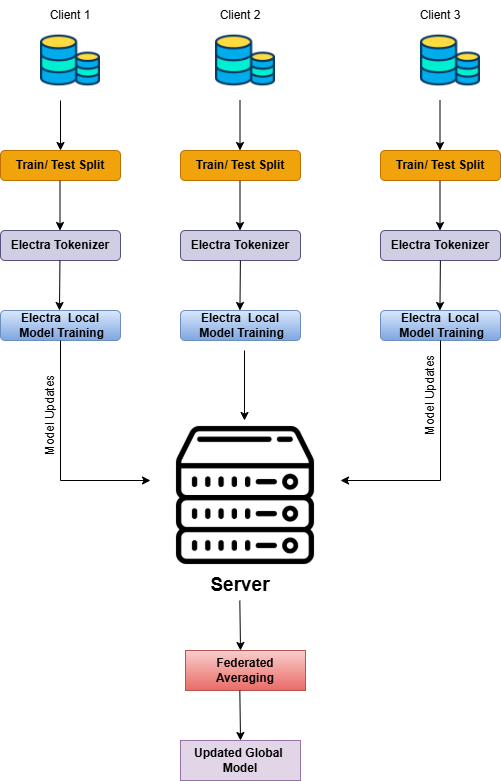}
\caption{Architecture of the proposed Federated ELECTRA model.}
\label{fig:federated_electra_architecture}
\end{figure}

Figure~\ref{fig:federated_electra_architecture} illustrates the architecture of the proposed Federated ELECTRA model. The workflow begins at the client level, where the distributed dataset is available across multiple clients. Each client performs local data preparation followed by train/test splitting and text tokenization using the ELECTRA tokenizer. The tokenized data are then used to train the local ELECTRA model on each client independently. After local training, only the model updates are transmitted to the central server while the raw client data remain locally stored. The server aggregates the received model parameters using the Federated Averaging (FedAvg) algorithm to generate an updated global model.

\subsection{Explainability}
To improve the interpretability of the sentiment classification model, we included Explainable Artificial Intelligence (XAI) in the analysis. In this case, the SHAP method has been employed to evaluate the effects that individual words have on the predictions generated by the model. This is achieved through the calculation of the contribution score for each word, which can be used to determine the words that improve the probability of the particular sentiment class being predicted \cite{apu2025explainable, aurpa2026transparent}.

\subsection{Proposed Methodology}

\begin{figure}[htbp]
\centering
\includegraphics[width=\textwidth]{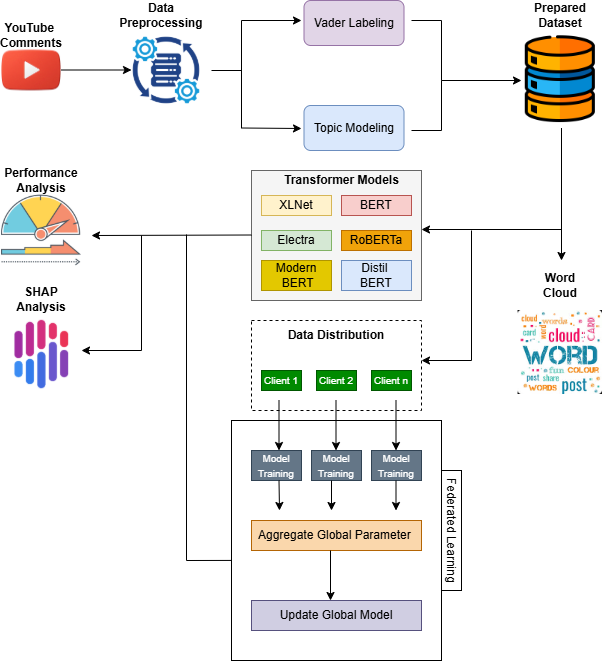}
\caption{Overall workflow of the proposed framework including data preprocessing, sentiment labeling, topic modeling, transformer-based model comparison, federated learning and explainable AI analysis.}
\label{fig:proposed_methodology}
\end{figure}

Figure~\ref{fig:proposed_methodology} presents the overall workflow of the proposed framework. The study begins with the collection of YouTube comments related to the war situation. The collected data are first subjected to preprocessing, including cleaning, normalization, and removal of irrelevant text elements in order to prepare the dataset for analysis.

After preprocessing, sentiment labels are generated using the VADER sentiment analyzer, while topic modeling is applied to identify major discussion themes within the comments. The labeled and processed dataset is then used to train different transformer-based models such as XLNet, BERT, ELECTRA, RoBERTa, ModernBERT and DistilBERT. These models are evaluated and compared to identify the best-performing architecture for sentiment classification.

Based on the performance comparison, ELECTRA was selected as the most effective model. The prepared dataset is then distributed across multiple clients to simulate a federated learning environment. Each client performs local model training independently on its portion of the data, after which the model parameters are transmitted to a central server. The server aggregates the local updates using the Federated Averaging (FedAvg) algorithm to generate an updated global model.

In the end, the performance of the federated model is evaluated using standard classification metrics. To improve model transparency and interpretability, XAI techniques based on SHAP are applied to analyze how individual words influence the sentiment predictions produced by the ELECTRA model.
\section{Results}\label{sec3}

This section presents the results and analysis of the proposed methodology. The results include the exploratory analysis of the gathered comments, the topic-wise sentiment analysis, the performance analysis of the sentiment classification models using the transformer architecture, and the performance analysis of the proposed federated ELECTRA model for sentiment classification with varying settings of the clients.

\subsection{Exploratory Analysis of Public Discourse}
To better understand the nature of discussions surrounding the USA–Iran conflict, an exploratory analysis was conducted on the collected YouTube comments. This analysis aims to identify dominant themes, sentiment patterns, and commonly used words in the public discourse. By applying topic modeling and visualization techniques, such as LDA topic extraction and word cloud analysis, we explore how users discuss different aspects of the conflict, including political narratives, military actions, and global geopolitical concerns. These insights provide an initial overview of the key topics and sentiment trends present in the dataset before conducting deeper sentiment classification and federated learning experiments.

\subsubsection{Topic Distribution Analysis of War-related Discussions}

\begin{figure}[htbp]
\centering
\includegraphics[width=0.85\textwidth]{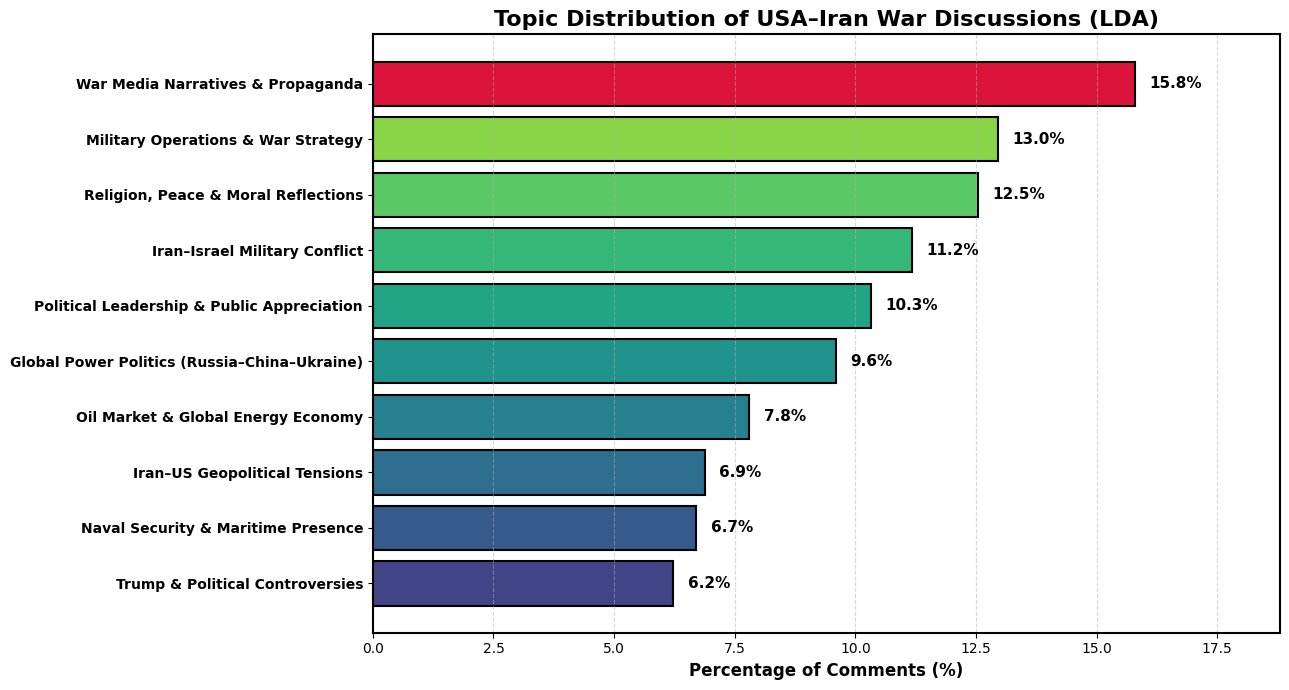}
\caption{Topic distribution of USA–Iran war related comments generated using LDA.}
\label{fig:topic_distribution}
\end{figure}

Figure~\ref{fig:topic_distribution} presents the distribution of the main topics identified from the YouTube comments related to the USA–Iran conflict using the LDA topic modeling technique. Topic modeling is commonly used in text mining to identify latent thematic structures in large bodies of textual data by identifying frequently co-occurring words into meaningful themes. 

The results reveal several dominant themes in public discussions. The most prominent topic is \textit{War Media Narratives and Propaganda}, which accounts for 15.8\% of the comments. This indicates that a significant portion of users are discussing media framing, misinformation, and narratives surrounding the conflict. The second most frequent topic is \textit{Military Operations and War Strategy} (13.0\%), reflecting discussions related to military actions, tactical decisions, and strategic developments in the conflict.

Another important topic is \textit{Religion, Peace and Moral Reflections} (12.5\%), where users express ethical concerns, religious viewpoints, and calls for peace. Discussions related to the \textit{Iran–Israel Military Conflict} (11.2\%) also appear prominently, indicating that many users connect the USA–Iran conflict with broader regional tensions in the Middle East.
Additional themes include \textit{Political Leadership and Public Appreciation} (10.3\%), where comments focus on political leaders and their roles in the conflict, and \textit{Global Power Politics} (9.6\%), which reflects discussions about the involvement of global powers such as Russia, China, and Ukraine. Economic implications are also present through the topic \textit{Oil Market and Global Energy Economy} (7.8\%), highlighting concerns about energy supply and global oil prices.

Other notable topics include \textit{Iran–US Geopolitical Tensions} (6.9\%), \textit{Naval Security and Maritime Presence} (6.7\%), and \textit{Trump and Political Controversies} (6.2\%), showing that political debates and international security issues also play a role in shaping public discourse.

\subsubsection{Topic-wise Sentiment Dynamics}

Figure~\ref{fig:sentiment_topic_distribution} shows the distribution of negative, neutral, and positive sentiment for the identified topics of discussion. The results show that the sentiment of the public varies significantly based on the topic of discussion. Some topics elicit high emotional responses, especially those related to geopolitical conflicts and political controversies. The detailed sentiment distribution of each topic is as under:

\begin{figure}[htbp]
\centering
\includegraphics[width=0.95\textwidth]{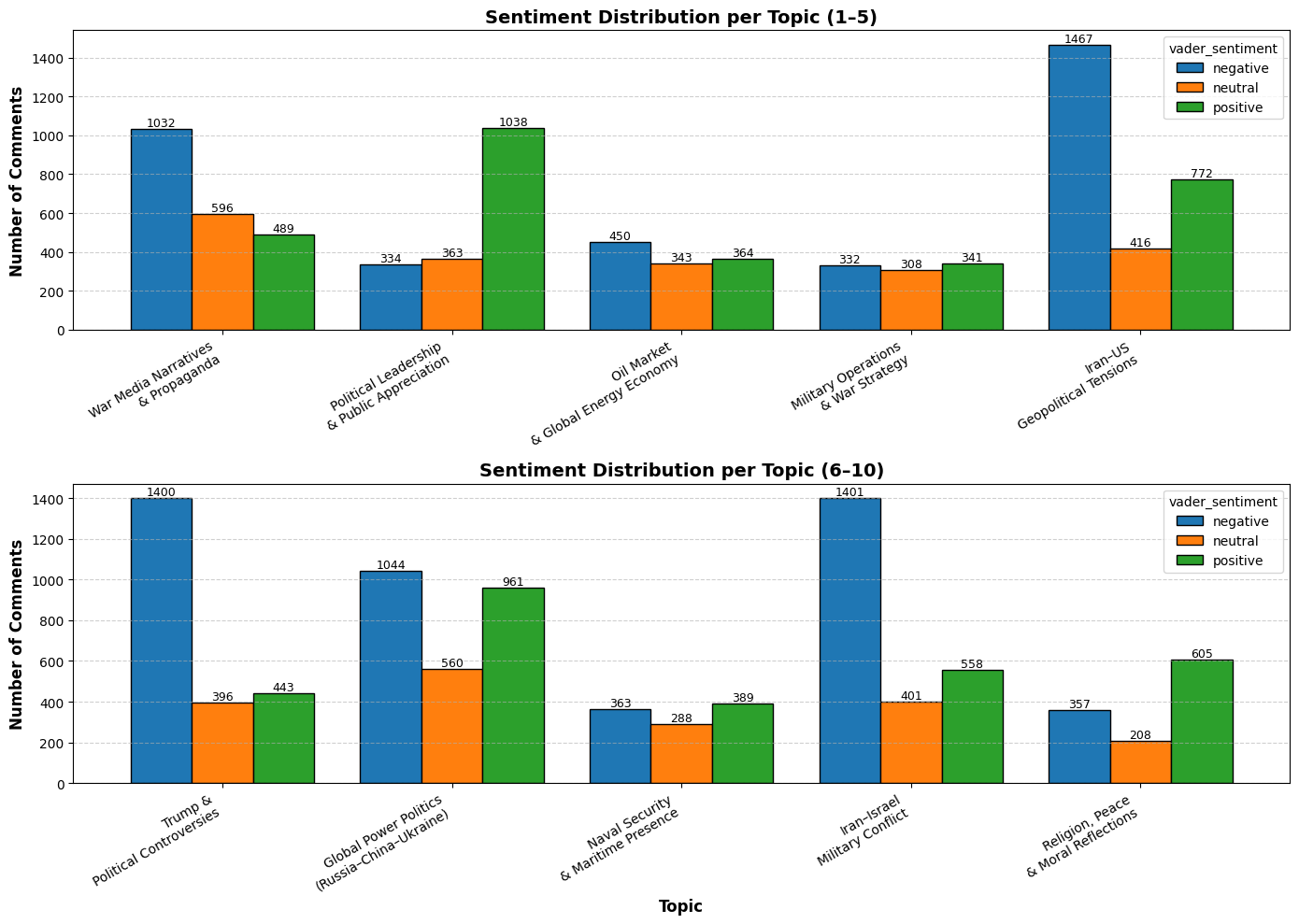}
\caption{Sentiment distribution across different topics showing the number of negative, neutral, and positive comments.}
\label{fig:sentiment_topic_distribution}
\end{figure}

\begin{itemize}

\item \textbf{War Media Narratives \& Propaganda:} The number of negative comments in this topic is 1032, while the number of neutral comments is 596, and positive comments is 489. The proportion of such negative responses is quite high, which may indicate that a significant portion of the users distrusted or even criticized the role of media narratives and propaganda in the creation of impressions about the conflict. There are frequent debates and accusations of misinformation, biased news coverage and mistrust in media institutions.

\item \textbf{Political Leadership \& Public Appreciation:}  In this topic, positive comments are more than others, with 1038 positive comments, while negative comments are 334 and neutral comments are 363.  It shows that there is a significant percentage of the users who show their support or appreciation towards some political figures. Many comments praise potential future leadership following discussions about the possible death of Iran’s Supreme Leader, Ayatollah Khamenei, while some users also express favorable opinions toward the role of the United States in the geopolitical situation. Many comments highlight leadership decisions, diplomatic actions or perceived strengths of political figures involved in the geopolitical situation.

\item \textbf{Oil Market \& Global Energy Economy:} The sentiment of this topic is relatively balanced, with 450 negative comments, 343 neutral comments, and 364 positive comments.  These discussions often focus on the economic consequences of the conflict, such as rising oil prices, energy supply concerns and the potential impact on global markets.  Many of these concerns are linked to discussions about the possible blockage of the Strait of Hormuz, a critical maritime route for global oil transportation.
The balanced sentiment suggests that users discuss these issues from both critical and analytical perspectives.

\item \textbf{Military Operations \& War Strategy:} The sentiment of this topic is relatively balanced, with 332 negative comments, 308 neutral comments, and 341 positive comments.  A large number of users speak of military tactics, defense strategy and operational developments in a more analytical or informative manner. As a result, the sentiment appears less polarized compared to more politically sensitive topics.

\item \textbf{Iran--US Geopolitical Tensions:} A strong negative sentiment is observed in this topic, with 1467 negative comments compared to 416 neutral and 772 positive comments. The strong sentiment can be attributed to the nature and content of the discussions and debates regarding international relations and tensions between Iran and the US.  Many comments express concern, frustration or disagreement regarding international policies and military actions.

\item \textbf{Trump \& Political Controversies:} Discussions related to this topic contain 1400 negative comments, 396 neutral comments, and 443 positive comments. Political controversies often trigger strong emotional reactions among users. Many comments reflect political disagreements, criticism of leadership decisions and polarized opinions about the role of political figures in international conflicts.

\item \textbf{Global Power Politics (Russia--China--Ukraine):} The number of comments under this topic is 1044 negative, 560 neutral, and 961 positive. The significant number of negative and positive comments indicates that users have different views about the involvement of the world’s powerful countries.

\item \textbf{Naval Security \& Maritime Presence:} The sentiment distribution consisting with 363 negative comments, 288 neutral comments and 389 positive comments. Some comments specifically refer to reported incidents involving U.S. military actions against vessels associated with Iran, including discussions about an Iranian-linked ship that was allegedly attacked while returning from naval exercises conducted near the Indian Ocean region. There are discussions on the Iranian ship that was allegedly attacked while on its way back from the naval exercises carried out near the Indian Ocean region. Due to the strategic movements associated with the incidents, the overall sentiment is relatively balanced.

\item \textbf{Iran--Israel Military Conflict:} This topic also shows dominant negative sentiment with 1401 negative comments, compared to 401 neutral and 558 positive comments.  The reflection of high negative sentiments towards the conflicts and regional military tensions. Many user express concern about the escalation, security threats and broader implications for regional stability.

\item \textbf{Religion, Peace \& Moral Reflections:} The topic presents 605 positive comments, 357 negative comments and 208 neutral comments. The topic generally focuses on the moral and religious aspects and peace, which is the reason it has the highest number of positive comments.

\end{itemize}

On a broader scale, it can be said from the above topic-wise sentiment analysis that discussions on geopolitical issues or political controversies often lead to higher negative sentiments, while discussions on appreciating political leadership, religion or peace often lead to more positive or neutral discussions. All these aspects clearly indicate how different aspects of international conflict affect public emotions.

\subsubsection{Word Cloud Based Sentiment Analysis}

The visualization of words cloud was applied to investigate the most common words that appeared in comments of various sentiment categories.  The identified methodology allows presenting the picture of the most popular words describing the view of the online debates concerning the conflict between the USA and Iran. By examining the linguistic patterns across positive, negative and neutral comments, it becomes possible to better understand the emotional tone, thematic focus and underlying concerns expressed through users.

\textbf{Positive Sentiment Word Cloud:}

\begin{figure}[H]
\centering
\includegraphics[width=0.75\textwidth]{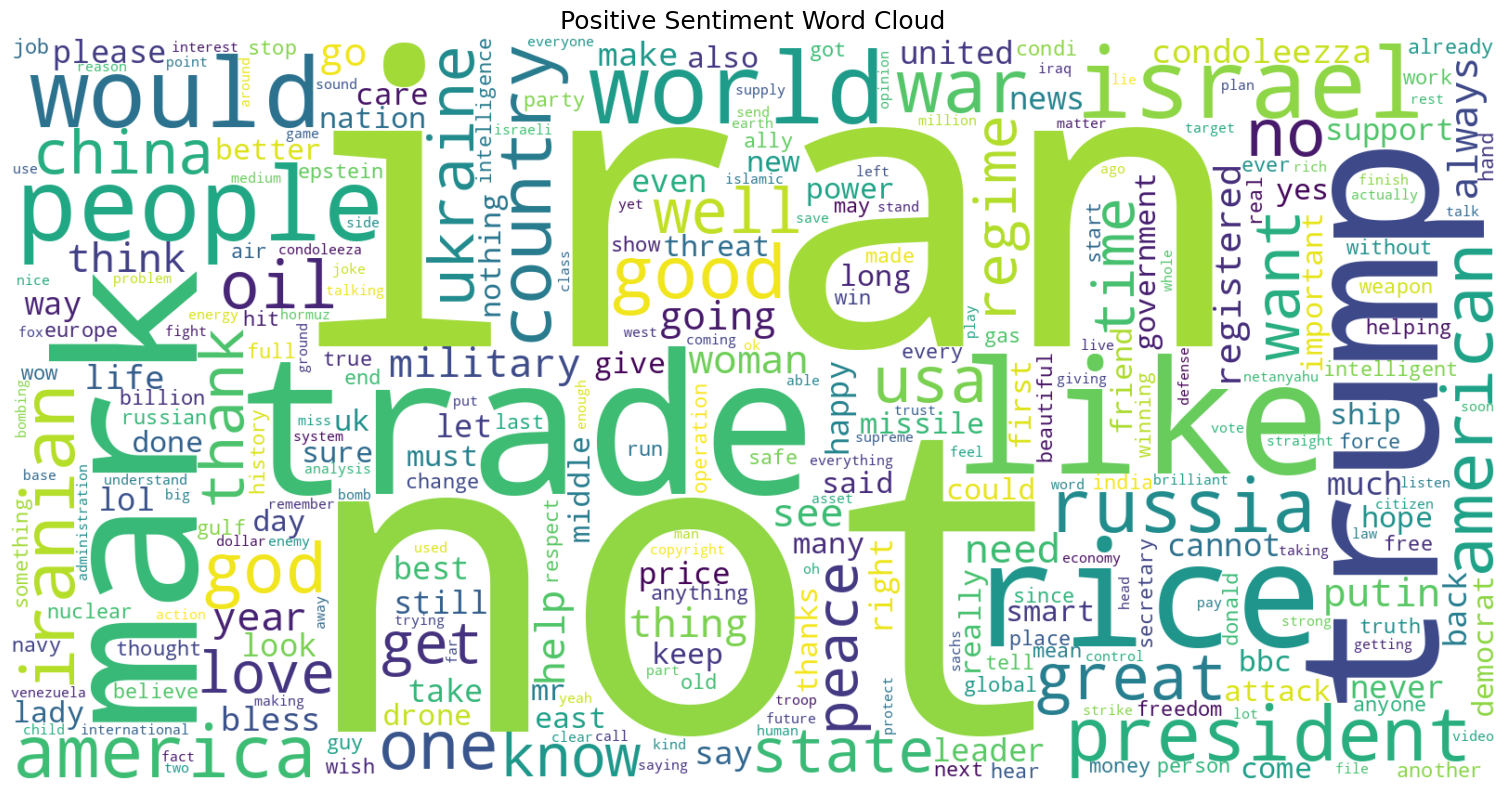}
\caption{Word cloud visualization representing the most frequent words appearing in positive sentiment comments.}
\label{fig:positive_wordcloud}
\end{figure}

Figure~\ref{fig:positive_wordcloud} illustrates the most frequent words appearing in comments classified as positive sentiment. Prominent terms include \textit{iran}, \textit{people}, \textit{america}, \textit{israel}, \textit{good}, \textit{peace}, \textit{great}, \textit{god}, \textit{state}, and \textit{president}. he presence of these words can be interpreted as evidence that positive words tend to be supportive, appreciative, or optimistic about political leadership, identity of nations, or the potential of peaceful solutions . Words such as \textit{peace}, \textit{good}, and \textit{great} indicate a constructive tone in which users highlight cooperation, stability or favorable political decisions. 

Additionally, the presence of religious and moral references such as \textit{god} suggests that some positive expressions are framed through ethical or spiritual viewpoints. Overall, the positive sentiment comments tend to reflect encouragement, trust in leadership or optimism regarding political and diplomatic developments.

\begin{figure}[H]
\centering
\includegraphics[width=0.75\textwidth]{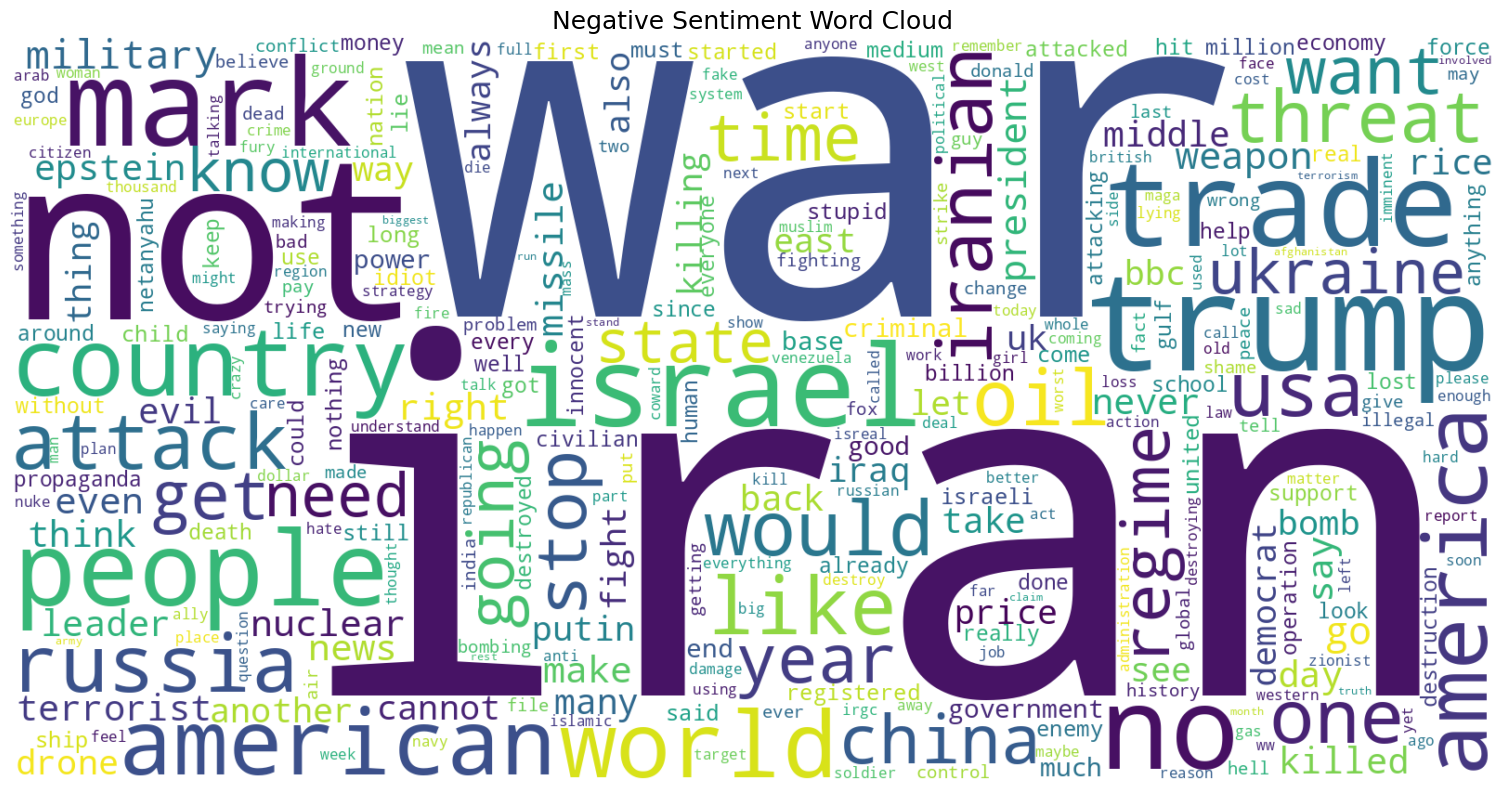}
\caption{Word cloud visualization showing the most common words appearing in negative sentiment comments.}
\label{fig:negative_wordcloud}
\end{figure}

\textbf{Negative Sentiment Word Cloud:}  
Figure~\ref{fig:negative_wordcloud} presents the dominant terms within negative sentiment comments. Frequently appearing words include \textit{war}, \textit{iran}, \textit{israel}, \textit{america}, \textit{russia}, \textit{attack}, \textit{fight}, \textit{bomb}, \textit{threat}, \textit{trump}, \textit{killing}, \textit{missiles}, \textit{epstein}, and \textit{nuclear}. These terms highlight the strong association between negative sentiment and discussions surrounding military conflict, geopolitical tension, and political controversy. Words such as \textit{war}, \textit{attack}, and \textit{missiles} reflect concerns related to armed confrontation and security threats, while references to political figures such as \textit{trump} indicate that leadership decisions also influence negative public reactions. These words indicate the high correlation between negative attitude and debates about military conflict, geopolitical tension, and political controversy. This pattern indicates that geopolitical discussions in online platforms are strongly shaped by perceptions of conflict and instability.

\textbf{Neutral Sentiment Word Cloud:}  

\begin{figure}[H]
\centering
\includegraphics[width=0.75\textwidth]{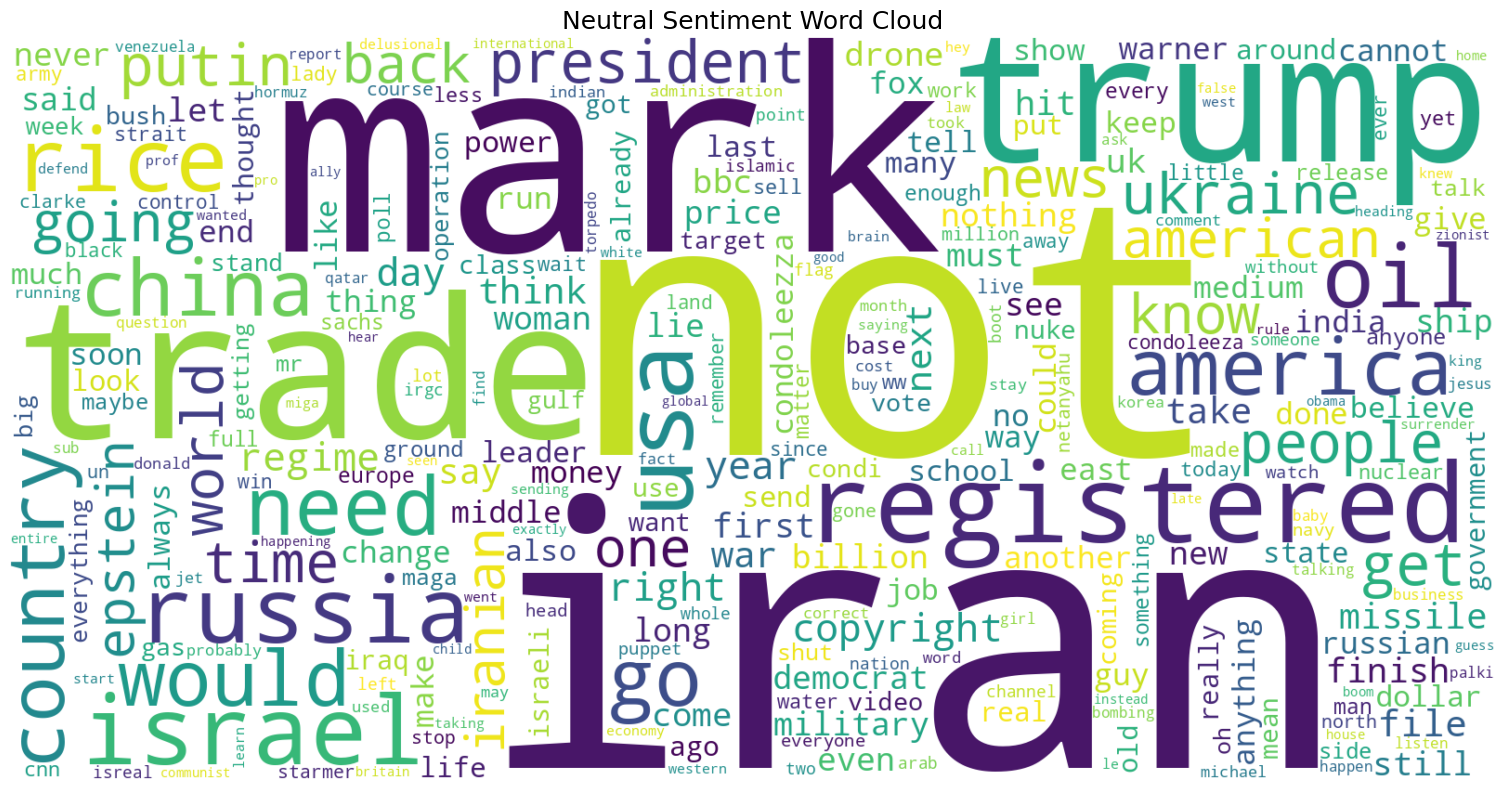}
\caption{Word cloud visualization of frequently used words in neutral sentiment comments.}
\label{fig:neutral_wordcloud}
\end{figure}
Figure~\ref{fig:neutral_wordcloud} displays the most frequent words appearing in neutral comments. Prominent terms include \textit{iran}, \textit{trade}, \textit{registered}, \textit{president}, \textit{america}, \textit{china}, \textit{oil}, and \textit{news}. These words indicate that neutral comments are generally more descriptive or informational rather than emotionally expressive. Numerous of these words can be linked to the geopolitical analysis, economic issues, or the discussion of the news.  For instance, words such as \textit{oil}, \textit{trade}, and \textit{china} suggest that users frequently discuss economic implications and international relations in a relatively objective manner. Similarly, terms like \textit{news} and \textit{report} indicate that some comments simply share information or observations without strongly positive or negative opinions. Therefore, neutral sentiment comments often represent analytical or factual discussions rather than emotionally driven reactions.

In general, the analysis of the word cloud shows a vivid image of the variation of the linguistic patterns between the categories of sentences of various sentiments. The conflict-related terms and expressions of geopolitical tension prevail in most negative comments, and supportive or optimistic terms related to the idea of leadership, peace, and national identity are most common in positive ones. Contrary to this, neutral comments are mostly concerned with informational or analytical discourses with regard to world politics and economic aspects. These results suggest that the emotional responses to the international conflicts, opinions, and the exchange of information are combined in the mass media discussions about social media networks.

\subsection{Sentiment Classification Performance}

Table \ref{tab:transformer_results} shows the performance comparison of different transformer models for sentiment classification. The performance of the models is compared using three different metrics  Accuracy, Macro F1 and Weighted F1 Score. These metrics are commonly used for performance evaluation of sentiment classification models.

As shown in the results, all transformer models have shown promising results for sentiment classification, with accuracy greater than 90\%. This shows that transformer models are highly efficient for sentiment analysis in social media comments.

\begin{table}[htbp]
\caption{Comparative performance of transformer-based models for sentiment classification}
\label{tab:transformer_results}
\centering
\begin{tabular}{lccc}
\hline
Model & Accuracy (\%) & Macro F1 (\%) & Weighted F1 (\%) \\
\hline
BERT        & 90.68 & 90.43 & 90.67 \\
RoBERTa     & 90.23 & 89.97 & 90.22 \\
XLNet       & 90.73 & 90.43 & 90.72 \\
DistilBERT  & 90.37 & 90.22 & 90.36 \\
ELECTRA     & \textbf{91.32} & \textbf{90.96} & \textbf{91.29} \\
ModernBERT  & 90.84 & 90.55 & 90.84 \\
\hline
\end{tabular}
\end{table}

Among the evaluated models, ELECTRA achieved the best overall performance, obtaining an accuracy of 91.32\%, a Macro F1-score of 90.96\%, and a Weighted F1-score of 91.29\%.  This shows that ELECTRA is slightly more effective in determining the sentiment polarity of the sentences in the dataset compared to the performance of the other models. The effectiveness of ELECTRA can be attributed to its efficient pre-training approach, which helps the model learn the contextual representation more efficiently.

ModernBert, XLNet and BERT are also other models that competed with competitive results with an accuracy exceeding 90\%, which can be interpreted as the subsequent models also being competent in sentiment classification tasks. The performance of the DistilBERT and RoBERTa models is slightly low compared to the other models, showing the effectiveness of transformer models in performing sentiment classification.

Overall, the comparative results indicate that while all evaluated transformer models perform well on the sentiment classification task, ELECTRA provides the most consistent and reliable performance among them. Therefore, ELECTRA was selected as the primary model for the subsequent experiments and analysis in this study.

\subsubsection{Evaluation of the Centralized Sentiment Classification Model}

\begin{figure}[htbp]
\centering
\includegraphics[width=0.8\textwidth]{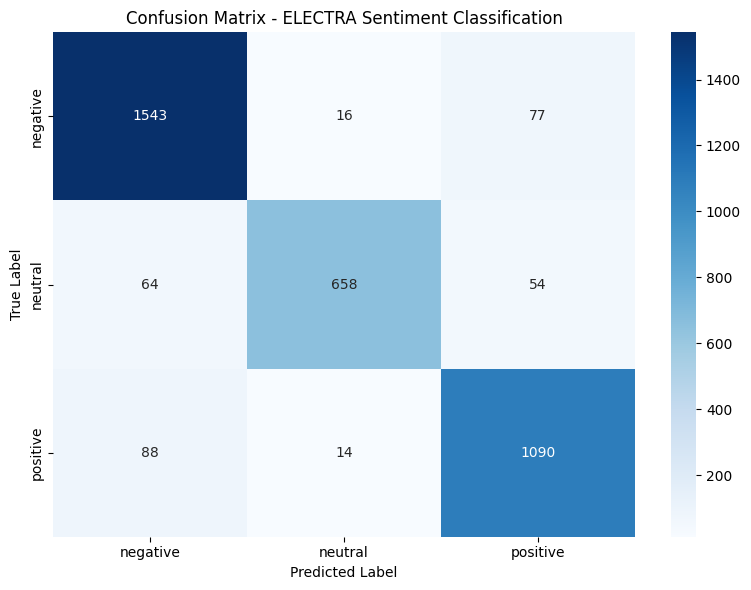}
\caption{Confusion matrix of the ELECTRA-based sentiment classification model.}
\label{fig:electra_confusion}
\end{figure}

Figure~\ref{fig:electra_confusion} presents the confusion matrix of the ELECTRA-based sentiment classification model on the test dataset. The confusion matrix represents the effectiveness of the model in classifying the three sentiment classes.

As indicated by the results, the model was able to classify a substantial number of comments in all the sentiment classes. In the case of the negative class, the model was able to correctly classify 1543 comments, while a small number of comments were misclassified, 16 as neutral and 77 as positive. In the case of the positive class, the model was able to correctly classify 1090 comments, while a small number of comments were misclassified, 88 as negative and 14 as neutral. In the case of the neutral class, the model was able to correctly classify 658 comments with a few misclassifications like 64 comments were predicted as negative and 54 comments were predicted as positives.

Based on the confusion matrix, it can be determined that the ELECTRA model is efficient in distinguishing different sentiment categories. The majority of predictions are along the diagonal line of the confusion matrix, which represents correct classification, and only a few comments are incorrectly classified. These results further validate the effectiveness of the ELECTRA model for sentiment analysis of the collected YouTube comment dataset.

\subsection{Federated Learning Performance under Different Client Settings}
Table~\ref{tab:fed_client_performance} presents the performance of the ELECTRA-based federated learning model across different client settings. In our experiment, the data were divided among several clients and the model was trained with a federated learning system where each client trained the model locally and then exchanged model updates with the central server.

\begin{table}[htbp]
\caption{Performance of the ELECTRA-based federated learning model under different client configurations}
\label{tab:fed_client_performance}
\centering
\begin{tabular}{lccc}
\hline
Number of Clients & Accuracy (\%) & Macro F1 (\%) & Weighted F1 (\%) \\
\hline
2 Clients & \textbf{89.59} & \textbf{89.30} & \textbf{89.57} \\
4 Clients & 86.88 & 86.23 & 86.79 \\
6 Clients & 85.27 & 84.29 & 85.14 \\
\hline
\end{tabular}
\end{table}

The findings indicate that a two client configuration was the most successful with an accuracy of 89.59\% and Macro F1-score of 89.30\% and a Weighted F1-score of 89.57\%.  However, as the number of clients increases to four and six, there is a gradual decrease in the performance of the system.

The reduction in the performance level may be due to the splitting of the dataset into multiple clients, which could affect the capacity of the model to learn all the patterns during the training phase. However, the federated learning model has good performance across all the configurations.

Overall, the results indicate that the ELECTRA-based federated learning framework can effectively perform sentiment classification while enabling distributed training across multiple clients.

\subsubsection{Confusion Matrix Analysis of Federated Learning Configurations}

\begin{figure*}[h]
\centering
\includegraphics[width= 0.8\textwidth]{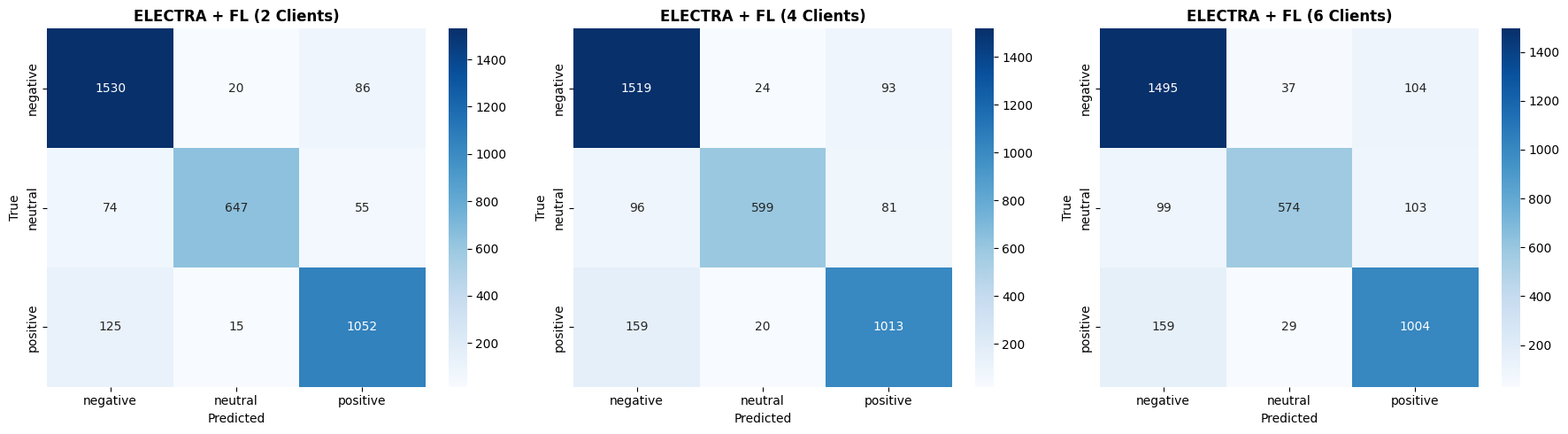}
\caption{Confusion matrixes of the ELECTRA-based federated learning model under different client configurations: (a) 2 clients, (b) 4 clients, and (c) 6 clients. The matrices illustrate the distribution of predicted versus true sentiment labels across the negative, neutral, and positive classes.}
\label{fig:federated_confusion}
\end{figure*}

Figure~\ref{fig:federated_confusion} presents the confusion matrix's of the ELECTRA-based federated learning model under different client configurations. Overall, the majority of predictions lie along the diagonal in all three matrices, indicating that the model correctly classifies most of the comments across the negative, neutral and positive sentiment classes.

In the two-client configuration, the model achieves the highest accuracy, correctly classifying 1530 negative, 647 neutral and 1052 positive comments. Only a small number of instances are misclassified, suggesting that the model learns sentiment patterns effectively when sufficient data are available per client.

As the number of clients increases to four and six, a slight increase in misclassification can be observed. For example, some negative comments are predicted as positive, and some neutral comments are classified as either negative or positive. This occurs mainly because the dataset is divided among more clients, reducing the amount of local training data available at each node.

Despite these minor errors, the overall structure of the matrices shows that the federated ELECTRA model maintains strong classification capability across all configurations. The results indicate that the model remains reliable for sentiment classification even in distributed training environments.

\subsection{Model Transparency Through Explainable AI Techniques}

In order to make the proposed sentiment classification model more interpretable, Explainable Artificial Intelligence (XAI) approaches were utilized in the testing phase of the model’s evaluation process. In this regard, SHAP (SHapley Additive exPlanations) is utilized to determine the influence of individual words on the model’s predictions. It is possible to determine which words are increasing or reducing the probability of a certain sentiment classification based on the SHAP values of individual words.

SHAP is utilized on a subset of the validation set for the purpose of explaining the predictions of the trained model in the evaluation phase of the model’s development process. In this regard, it is possible to determine the decision patterns learned by the complex transformer-based model without affecting the training process of the model. The magnitude of each SHAP value represents the strength of each word’s contribution to the predicted sentiment.

\begin{figure}[htbp]
\centering
\includegraphics[width=1.\textwidth]{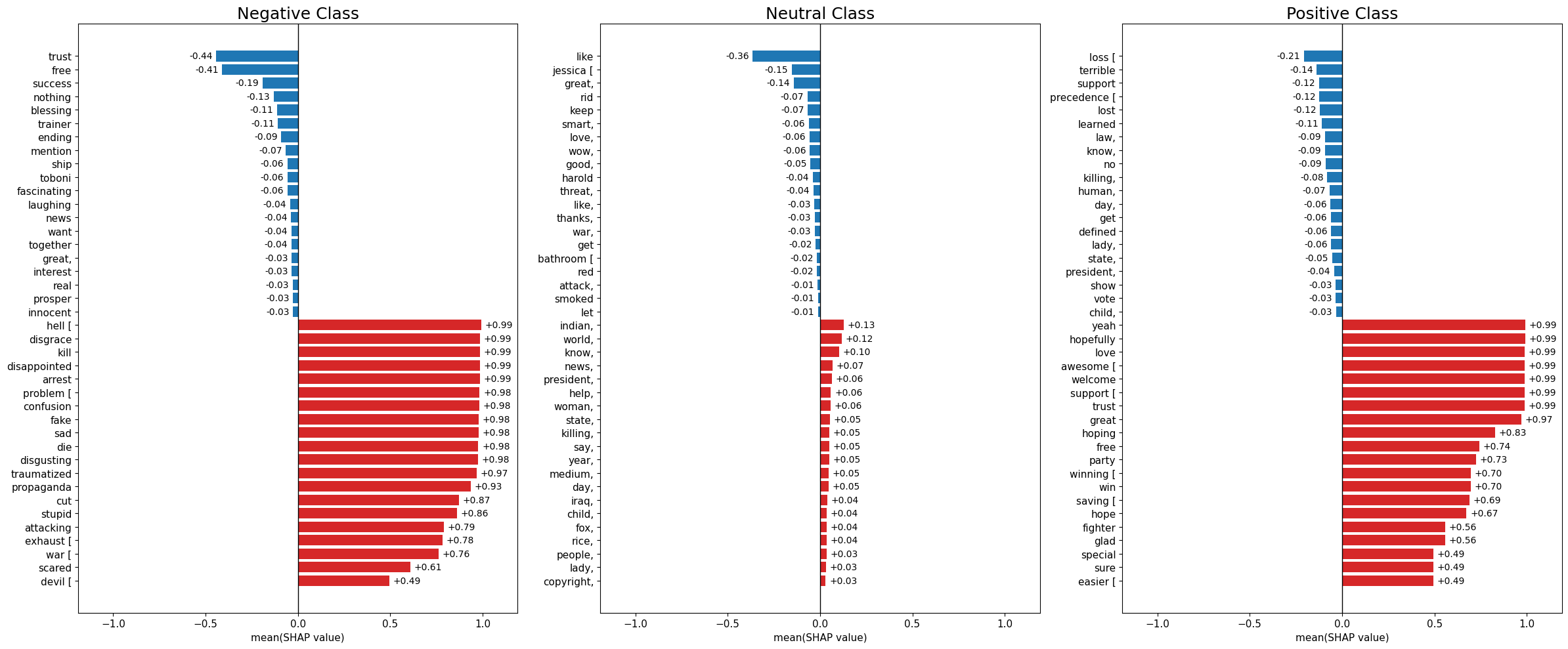}
\caption{Class-wise SHAP explanation showing the most influential tokens contributing to negative, neutral, and positive sentiment predictions. Red bars indicate tokens that increase the likelihood of the class, while blue bars represent tokens that decrease the probability of that class.}
\label{fig:shap_classwise_explanation}
\end{figure}

Figure~\ref{fig:shap_classwise_explanation} presents the most influential tokens for each sentiment class. For the \textbf{negative sentiment class}, words such as \textit{hell}, \textit{disgrace}, \textit{kill}, \textit{disappointed}, and \textit{sad} strongly increase the probability of negative predictions, while terms like \textit{trust} and \textit{success} reduce it. 

For the \textbf{neutral sentiment class}, influential words include \textit{world}, \textit{news}, \textit{president}, and \textit{state}, which typically appear in factual or descriptive discussions. 

For the \textbf{positive sentiment class}, words such as \textit{love}, \textit{awesome}, \textit{welcome}, and \textit{great} strongly contribute to positive predictions, whereas terms like \textit{loss} and \textit{terrible} decrease the likelihood of a positive classification.

Thus, the overall result of the SHAP analysis is that the model makes use of significant linguistic features in its prediction. The emotionally charged words contribute to the negative sentiment prediction, the descriptive words contribute to the neutral sentiment prediction, and the supportive words contribute to the positive sentiment prediction.

 \section{Discussion}

The experimental findings offer some ideas both about the nature of the public discourse of the USAIran conflict and the efficiency of the suggested sentiment analysis model. The exploratory analysis indicates that discussions about the conflict are distributed across several thematic dimensions. The largest topic identified through LDA is \textit{War Media Narratives and Propaganda}, accounting for 15.8\% of the comments, followed by \textit{Military Operations and War Strategy} (13.0\%) and \textit{Religion, Peace and Moral Reflections} (12.5\%). These results suggest that the field of online discussions is not limited to the military events only but also includes the media narratives and ethical issues and geopolitical discussions.

The sentiment analysis by topic displays a great deal of diversity in the emotional reaction based on the subject of discussion. For example, the topic \textit{Iran--US Geopolitical Tensions} contains the highest number of negative comments (1467), compared to 772 positive and 416 neutral comments. Similarly, the \textit{Trump and Political Controversies} topic shows strong negative sentiment with 1400 negative comments. In contrast, the topic \textit{Political Leadership and Public Appreciation} contains a higher proportion of positive comments (1038), suggesting that discussions about leadership often involve expressions of support or approval. These findings indicate that geopolitical issues and political scandals cause more negative responses, and the topic regarding leadership or moral considerations causes comparatively more positive responses.

The word cloud analysis also outlines unique linguistic patterns by sentiment. Positive comments frequently include words such as \textit{peace}, \textit{good}, \textit{great}, \textit{god} and \textit{people}, indicating supportive and optimistic expressions related to leadership, diplomacy or hopes for stability.  In contrast, negative comments are dominated by conflict-related terms such as \textit{war}, \textit{attack}, \textit{bomb}, \textit{threat}, \textit{missiles} and \textit{nuclear}, reflecting public concern about military escalation and geopolitical instability. Neutral comments contain more informational or analytical terms such as \textit{oil}, \textit{trade}, \textit{news} and \textit{china}, suggesting that many users discuss the conflict in a descriptive or analytical manner rather than expressing strong emotions.

The transformer-based classification experiments shows strong performance across all evaluated models.  All six transformer architectures achieved accuracy values above 90\%, confirming the suitability of transformer-based approaches for sentiment analysis of social media comments. Among the models, ELECTRA achieved the best overall performance with an accuracy of 91.32\%, a Macro F1-score of 90.96\% and a Weighted F1-score of 91.29\%. Other models such as ModernBERT (90.84\%), XLNet (90.73\%) and BERT (90.68\%) also produced competitive results. The effectiveness of ELECTRA can also be explained by its replaced-token detection pretraining method, which enables the model to acquire contextual relationships more effectively than the conventional masked language models.

The confusion matrix results further confirm the effectiveness of the centralized ELECTRA model. The model correctly classified 1543 negative comments, 1090 positive comments, and 658 neutral comments, with only a limited number of misclassifications. This indicates that the model successfully captures contextual cues associated with different sentiment classes.

The experiments carried out in federated learning offer further insights into the trade-off between distributed training and classification performance. The two-client configuration achieved the best performance with an accuracy of 89.59\%, Macro F1-score of 89.30\%, and Weighted F1-score of 89.57\%. However, as the number of clients increased to four and six, the performance decreased to 86.88\% and 85.27\% accuracy, respectively. This is attributed to data fragmentation, which limits the amount of data for training models.

Finally, the explainability analysis using SHAP provides valuable insights into the model's decision-making process. The SHAP results show that emotionally negative words such as \textit{hell}, \textit{disgrace}, \textit{kill}, \textit{disappointed} and \textit{sad} strongly increase the probability of negative sentiment predictions. In contrast, positive tokens such as \textit{love}, \textit{awesome}, \textit{welcome}, \textit{trust}, and \textit{great} significantly increase the probability of positive sentiment classification. Words such as \textit{world}, \textit{news}, \textit{president}, and \textit{state} appear as influential features in the neutral class, indicating their role in descriptive or factual discussions. These explainability results confirm that the model relies on meaningful linguistic patterns when making predictions, which increases the transparency and reliability of the proposed framework.

Overall, topic modeling combined with sentiment classification with the help of the transformer architecture, federated learning to preserve privacy and explainable AI provides a complete solution to the problem of the analysis of the public discussion in the context of international conflicts. The findings have shown that the suggested framework is not solely useful in terms of prediction but also offers understanding of the underlying factors that can affect the public sentiment on social media.

\section{Conclusion}\label{sec13}

This study presented a framework for analyzing public discourse related to the USA--Iran conflict by combining topic modeling, transformer-based sentiment classification, federated learning, and explainable AI. The exploratory analysis revealed that the key themes that are discussed predominantly are the narratives of war, geopolitical tensions, political leadership, and moral considerations. The sentiment analysis also showed that the geopolitical conflicts and the political scandals disposition are more likely to produce more negative responses and the issues concerning leadership and moral values receive comparatively more positive reponses.

Among the evaluated transformer models, ELECTRA achieved the best performance with an accuracy of 91.32\%, a Macro F1-score of 90.96\%, and a Weighted F1-score of 91.29\%. The federated ELECTRA framework also demonstrated that sentiment classification can be performed effectively in a distributed setting while preserving data privacy, achieving 89.59\% accuracy in the two-client configuration. This privacy-preserving approach allows models to learn from decentralized data without sharing raw user information.

However, some limitations remain. The sentiment labels were generated automatically. In addition, the federated experiments were conducted in a simulated environment with a limited number of clients and the analysis focused only on YouTube comments.

Future studies can examined on manual annotated datasets, combining multimodal data sources to additionally enhance the preciseness and practicability of privacy-conscious sentiment analysis systems.

\section*{Data and Code Availability}
The code and data used in this paper are available in the following GitHub repository. 
https://github.com/msijewelsaif/USA-IRAN-Conflict-2026

\bibliography{sn-bibliography}

\end{document}